\PassOptionsToPackage{table,xcdraw}{xcolor}
\documentclass[sigconf]{acmart}
\usepackage{multirow}
\usepackage{bbding}
\usepackage{subfig} 
\usepackage{graphicx}   
\usepackage{subfig}     
\usepackage{caption}    
\usepackage{booktabs}   
\usepackage{array}
\usepackage[table]{xcolor}
\AtBeginDocument{%
  }

\setcopyright{acmlicensed}
\copyrightyear{2018}
\acmYear{2018}
\acmDOI{XXXXXXX.XXXXXXX}
\acmConference[MM '25]{Proceedings of the 33th ACM International Conference on Multimedia}{October 27--31, 2025}{Dublin, Ireland}
\acmISBN{978-1-4503-XXXX-X/2018/06}

\begin{document}

\title{FingerVeinSyn-5M: A Million-Scale Dataset and Benchmark for Finger Vein Recognition}

\author{Yifan Wang}
\affiliation{%
	\institution{School of Cyber Science and
		Engineering, Southeast University}
	\city{Nnajing}
	\country{China}
}
\email{230239767@seu.edu.cn}
\author{Jie Gui}
\authornote{Corresponding author}
\affiliation{%
	\institution{School of Cyber Science and
		Engineering, Southeast University}
	\city{Nanjing}
	\country{China}
}
\email{guijie@seu.edu.cn}

\author{Baosheng Yu}
\affiliation{%
	\institution{Lee Kong Chian School of Medicine, Nanyang, Technological University}
	\country{Singapore}}
\email{aosheng.yu@ntu.edu.sg}

\author{Qi Li}
\affiliation{%
	\institution{Institute of Automation, Chinese Academy of Sciences}
	\city{Beijing}
	\country{China}}
\email{qli@nlpr.ia.ac.cn}

\author{Zhenan Sun}
\affiliation{%
	\institution{Institute of Automation, Chinese Academy of Sciences}
	\city{Beijing}
	\country{China}
}
\email{znsun@nlpr.ia.ac.cn}
\author{Juho Kannala}
\affiliation{%
	\institution{Department of Computer Sccience, Aalto University}
	\city{Helsinki}
	\country{Finland}}
\email{Juho.Kannala@aalto.fi}

\author{Guoying Zhao}
\affiliation{%
	\institution{Center for Machine Vision and Signal Analysis , University of Oulu}
	\city{Oulu}
	\country{Finland}}
\email{guoying.zhao@oulu.fi}

\renewcommand{\shortauthors}{Yifan Wang et al.}

\begin{abstract}
A major challenge in finger vein recognition is the lack of large-scale public datasets. Existing datasets contain few identities and limited samples per finger, restricting the advancement of deep learning-based methods. To address this, we introduce FVeinSyn, a synthetic generator capable of producing diverse finger vein patterns with rich intra-class variations. Using FVeinSyn, we created FingerVeinSyn-5M -- the largest available finger vein dataset --  containing 5 million samples from 50,000 unique fingers, each with 100 variations including shift, rotation, scale, roll, varying exposure levels, skin scattering blur, optical blur, and motion blur. FingerVeinSyn-5M is also the first to offer fully annotated finger vein images, supporting deep learning applications in this field. Models pretrained on FingerVeinSyn-5M and fine-tuned with minimal real data achieve an average 53.91\% performance gain across multiple benchmarks. The dataset is publicly available at: \url{https://github.com/EvanWang98/FingerVeinSyn-5M}.
\end{abstract}




\maketitle

\section{Introduction}
Finger vein recognition (FVR) has made significant progress over the past two decades. Automated FVR systems are increasingly deployed, with companies such as Hitachi, Fujitsu, and Tencent introducing commercial products. While mature biometrics like face~\cite{9763004, 10.1145/3664647.3680704, 10.1145/3664647.3680635} and fingerprint~\cite{9893541} recognition are widely used, they raise concerns regarding security and privacy. In contrast, finger vein recognition is difficult to spoof, offers stable uniqueness due to internal vascular patterns, and does not rely on surface features—providing stronger privacy protection and reducing the risk of misuse~\cite{10620353}.

\begin{table}[!t]
	\centering
		\renewcommand{\arraystretch}{0.4}
	\caption{Comparison of the scales of publicly available face, fingerprint, and finger vein datasets.}
	\label{tb1}
	\begin{tabular}{lcc}
		\toprule
		Dataset            & \#IDs & \#Images   \\ \midrule
		\rowcolor{gray!20}\multicolumn{3}{l}{\textbf{Face}}                 \\ \midrule
		VGGFace2~\cite{cao2018vggface2}            & 9131  & 3.3M \\
		MS-Celeb~\cite{guo2016ms}            & 100k  & 10M  \\
		WebFace260M~\cite{9763004}         & 4M    & 260M       \\ \midrule
		\rowcolor{gray!20}\multicolumn{3}{l}{\textbf{Fingerprint}}          \\ \midrule
		NIST SD 302~\cite{nist302}         & 2000  & 25093      \\
		CASIA-FingerprintV5~\cite{bitdataset} & 4000  & 20000      \\
		NIST SD 300~\cite{fiumara2018nist}         & 8880  & 17760      \\ \midrule
		\rowcolor{gray!20}\multicolumn{3}{l}{\textbf{Finger Vein}}          \\ \midrule
		SDUMLA-FV~\cite{yin2011sdumla}           & 636   & 3816       \\
		SCUT-FV~\cite{tang2019finger}             & 696   & 8526       \\
		NUPT-FV~\cite{ren2022dataset}             & 860   & 16800      \\ \midrule
		FingerVeinSyn-5M (\textbf{Ours})			& 50k&  5M \\ \bottomrule
	\end{tabular}
\end{table}

Despite its advantages, the progress of FVR is limited by the lack of large-scale public datasets~\cite{WANG2023119874}. Unlike face recognition, which leverages abundant online images for training deep networks~\cite{9763004}, FVR—like fingerprint recognition—requires data collection through specialized hardware. This makes large-scale dataset creation time-consuming and resource-intensive. The scarcity of publicly available training data constrains the development of deep learning models and hampers the comprehensive evaluation of large-scale recognition performance, such as retrieval accuracy and efficiency at scale.

Currently available finger vein datasets include FV-USM~\cite{asaari2014fusion}, MMCBNU\_6000~\cite{lu2013available}, SDUMLA-FV~\cite{yin2011sdumla}, UTFVP~\cite{Ton_UTFVPDB2013}, VERA-FV~\cite{vanoni2014cross}, PLUSVein~\cite{8698588}, SCUT-FV~\cite{tang2019finger}, NUPT-FV~\cite{ren2022dataset} and HKPU-FV~\cite{kumar2011human}. However, they share key limitations: 1) \textbf{Limited number of unique identities}: The largest datasets contain fewer than 1,000 distinct finger vein identities; 2)  \textbf{Few samples per identity}: Most datasets include only 2–12 impressions per finger; 3) \textbf{Low intra-class variation}: Samples are typically collected in a single session, offering minimal variation; at most, two sessions are provided; and 4) \textbf{Lack of standardized annotations}: Critical details—such as joint cavity positions, finger ROI, and shape—are often missing, making it difficult to apply advanced deep learning methods, particularly in unconstrained or non-contact scenarios.

Table \ref{tb1} highlights the significant disparity in the scale of publicly available datasets for face, fingerprint, and finger vein recognition. The largest existing face dataset comprises over 4 million identities and 260 million images~\cite{9763004}, while the largest fingerprint dataset includes nearly 10,000 unique identities~\cite{fiumara2018nist}. The availability of such extensive data has enabled rapid advancements in face and fingerprint recognition. In contrast, research in FVR continues to lag behind due to the scarcity of comparable large-scale datasets. Currently, only a limited number of studies focus on synthesizing finger vein data. Although generative models—such as adversarial networks and diffusion-based techniques—have been successful in face and fingerprint synthesis, they have struggled to effectively synthesize finger vein images due to the lack of sufficient real training data. Moreover, existing synthetic finger vein datasets and generators suffer from notable limitations in both data diversity and realism, making them inadequate for robust training and evaluation. Consequently, they continue to face the same challenges as real-world datasets.

To address these limitations, we introduce FVeinSyn, a dedicated finger vein synthesis framework. FVeinSyn comprises three main components: a vein pattern identity generator, a vein image renderer, and an intra-class variation generator. The identity generator uses L-System simulations constrained by anatomical models of finger structures to synthesize a vast number of distinct vein patterns. The renderer adopts a cascade region-aware GAN, trained on limited real data, to produce realistic finger vein images. The intra-class variation generator simulates a wide range of realistic variations for each identity, including translation, rotation, scaling, rolling, under-/overexposure, skin scattering blur, motion blur, and optical blur.

Using FVeinSyn, we created FingerVeinSyn-5M, the largest synthetic finger vein dataset to date. It contains 5 million images spanning 50,000 unique fingers, with 100 impressions per finger. To facilitate further progress in unconstrained and contactless finger vein recognition, FingerVeinSyn-5M also provides comprehensive annotations—including vein patterns, finger shapes, joint cavity locations, and ROI masks—enabling supervised training for a wide range of downstream tasks.

The main contributions of this study are as follows:
\begin{itemize}
	\item We introduce \textbf{FVeinSyn}, a finger vein image generator that produces a diverse set of unique identities and effectively models intra-class variations.
	\item We construct and release \textbf{FingerVeinSyn-5M}, the largest finger vein dataset to date, comprising 50,000 identities with 100 samples each, incorporating various intra-class transformations and their combinations.
	\item We provide the first \textbf{comprehensive annotation} set for finger vein images, covering vein patterns, finger shapes, joint cavity positions, and finger vein ROIs, which supports supervised learning across various contactless finger vein recognition tasks.
\end{itemize}

\section{Related Work}

\subsection{Finger Vein Recognition}
Finger vein recognition technology initially employed manual feature extraction methods~\cite{miura2004feature, miura2007extraction}, relying on handcrafted pattern descriptors and matching algorithms to achieve vein pattern classification. Although these traditional methods demonstrated basic recognition capabilities in controlled environments, their performance was constrained by inherent limitations such as insufficient feature representation and poor environmental adaptability.
With the advancement of deep learning, end-to-end training methods based on DCNNs have significantly improved recognition accuracy and system robustness, driving a paradigm shift in the field~\cite{8395431, 8979362, 9363648, 9564038, 10023509, 10695117, 9408606}. However, in scenarios involving cross-domain recognition or extreme data scarcity, existing deep learning approaches still face performance bottlenecks. While researchers have attempted to address these issues through strategies like data augmentation~\cite{10530126} and few-shot learning~\cite{9973284}, the improvements remain limited.
The core challenge currently hindering the development of finger vein recognition technology lies in the mismatch between deep representation capability and data scale. The lack of large-scale, high-quality datasets makes it difficult for models to learn highly generalizable feature representations. Establishing a comprehensive data support has become a key solution to overcoming existing technical barriers.

\subsection{Finger Vein Synthesis}
In recent years, synthetic biometric data generation technology has emerged as a key area of research to enhance recognition performance. In finger vein recognition, synthetic data generation is categorized into intra-class and inter-class generation~\cite{salazar2023towards}. The former focuses on creating supplementary samples for known identities, while the latter involves generating new unique identities and their multimodal samples. Most research has concentrated on intra-class generation, with only two studies exploring inter-class generation.
\cite{hillerstrom2014generating} made the first attempt at synthetic finger vein modeling and generation, releasing a synthetic dataset containing 5,000 identities with 10 samples per identity. Although this dataset is smaller than mainstream face and fingerprint datasets and faces challenges like limited sample authenticity and intra-class diversity, it has been crucial for advancing finger vein generation research. \cite{ou2022gan} adopted an ROI block randomization strategy to generate new identities and introduced GANs for image rendering, However, no dataset was constructed.

Significant progress has been made in related fields like palm vein~\cite{shang2025pvtree} and palmprint~\cite{jin2024pce, Shen_2023_ICCV, jin2025diffpalm} generation. In contrast, finger vein image generation faces greater challenges: 1) accurately simulating the anatomical structure of real fingers, including regions like joint cavities, and 2) existing research mostly focuses on ROI generation, while synthesizing complete finger samples (with geometric features and lighting effects) would greatly benefit deep representation learning and multi-task recognition. Thus, creating large-scale, high-fidelity synthetic finger vein datasets remains a critical challenge.

\begin{figure}[!t]
	\includegraphics[width=1.0\linewidth]{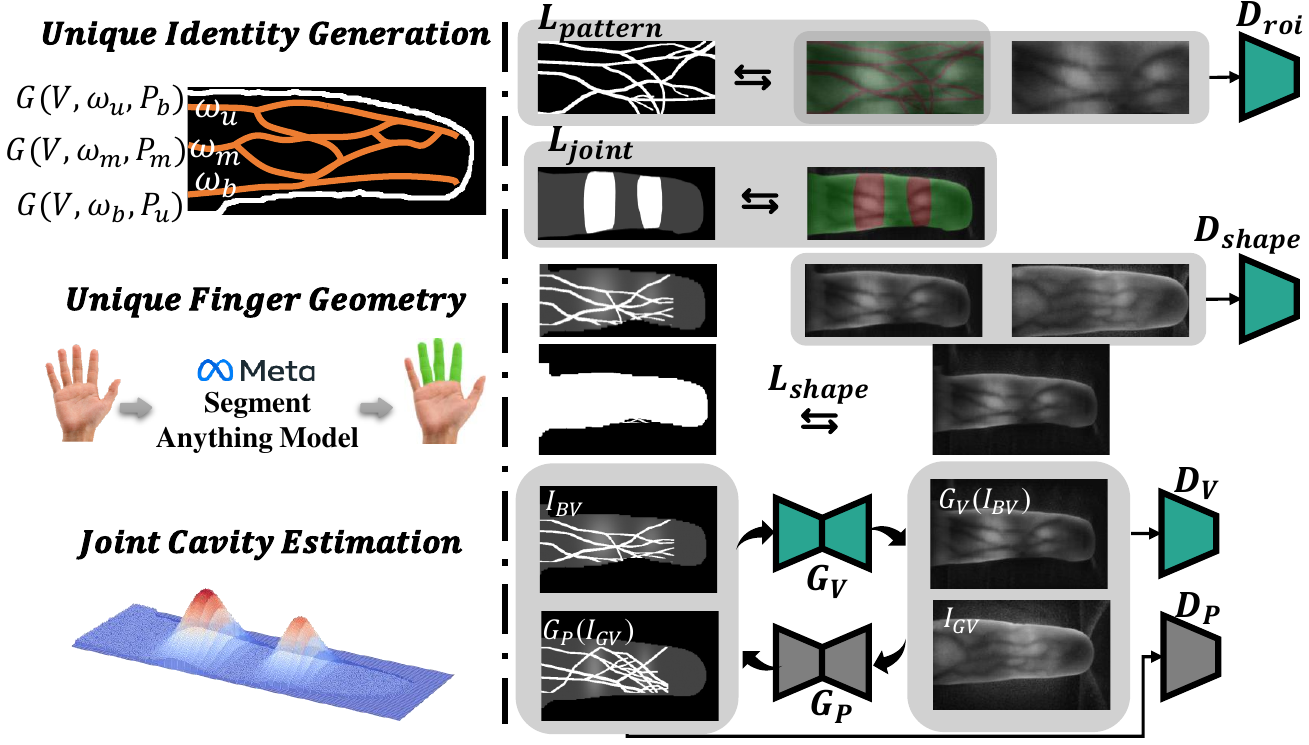}
	\caption{The main FVeinSyn framework for large-scale finger vein image synthesis. (Top) Unique pattern generation, estimation of finger shape and joint cavity regions are used for Finger Vein Pattern Generation. (Bottom) Cascaded region-aware GAN is used for Finger Vein Image Render.}
	\label{fig1}
\end{figure}

\section{Finger Vein Images Synthetic}

This section presents the FVeinSyn framework for large-scale finger vein image synthesis, illustrated in Fig.~\ref{fig1}. We start by introducing the L-system model for capturing complex finger vein branching patterns, followed by vascular network synthesis using stochastic L-Systems. Next, we describe the cascaded region-aware GAN for generating synthetic finger vein images from vein patterns. Lastly, we explain methods to enhance intra-class diversity.

\subsection{Preliminary}
The Lindenmayer System (L-System)~\cite{lindenmayer1968mathematical} is a formal grammar designed to describe the growth of simple multicellular organisms by iteratively replacing parts of an initial structure. It effectively models complex branching structures such as vascular networks in plants and animals. Formally, an L-System is defined as a tuple \( G = \{ V, \omega, P \} \), where \( V \) is a set of symbols (constants and variables),  \( \omega \) is the initial string from \( V \), and \( P \) is a set of production rules. For example, the Fibonacci L-System with \( V = \{a, b\} \), \( \omega = a \), and \( P = \{ a \rightarrow ab, b \rightarrow a \} \) generates the string \( abaababaabaababaababa \) after six iterations.  The stochastic L-System (S0L) extends this by assigning probabilities  \( p \) to rules, allowing nondeterministic rewriting. For instance,  \( P = \{ a: (p_i) \rightarrow ab, b: (p_j) \rightarrow a \} \) defines probabilistic productions for symbols a and b.

\subsection{Finger Vein Pattern Generation}
Finger vein patterns serve as highly distinctive biometric identifiers due to their intricate and unique structures. Each individual has a unique vein pattern distribution, with noticeable variations even between different fingers of the same person~\cite{wang2025colorvein}. Therefore, generating distinct and complex finger vein patterns is essential to creating reliable biometric identifiers. To ensure authenticity, synthesized vascular networks must comply with physiological constraints derived from real human anatomy. These include key morphological features such as vessel diameter, branch length, and spatial distribution, which collectively determine the uniqueness and biological plausibility of generated patterns~\cite{sweeney2024unsupervised}.

To this end, we propose a finger vein vascular network synthesis method based on stochastic L-Systems. This approach produces highly realistic vein pattern templates by controlling branch creation and segment deformation through grammatical rules. Specifically, we adapt the 3D vascular network generation framework (Vascular-System) from~\cite{sweeney2024unsupervised} into an optimized 2D solution by eliminating vessel twisting in 3D space and redefining the growth direction to strictly follow physiological patterns—extending from the finger base toward the fingertip. \citeauthor{hillerstrom2014generating} reveal that finger veins primarily consist of two main vessels on each side, connected by a venous network~\cite{hillerstrom2014generating}. Therefore, our synthetic vein pattern \( I_p \) is generated by:
\begin{equation}
\small
I_p = F_{\mathcal{B}}(F_{\mathcal{P}}(G_u(V, \omega_u, P_b), G_m(V, \omega_m, P_m), G_b(V, \omega_b, P_u)), T),
\end{equation}
where \( G(V, \omega, P) \) generates grammatical strings representing vein pattern, with \( \omega_u / \omega_m / \omega_b \) randomly assigning initial points to the upper, middle, and lower regions of the template \( T \) regions, respectively. The branching direction probabilities (up, down, forward) are controlled by \( P_u / P_m / P_b \). Middle veins branching from main vessels are randomly set to either one or two branches.  The function \( F_{\mathcal{P}} \) parses the grammar to output coordinates, while \( F_{\mathcal{B}} \) applies Bézier interpolation to smooth the veins and render them onto  \( T \), producing the binarized vein pattern \( I_p \) as a unique vein identity.

\subsection{Finger Vein Image Generation}

We design a Cascaded Region-Aware GAN (CascadedRA GAN) to generate synthetic finger vein images by rendering binary vein patterns \(I_p \) into full images. Inputs include \(I_p \), real finger shape \(I_s \), brightness distribution \(D_j \), and detailed annotations such as joint cavity coordinates, regions of interest, and finger variations. Instead of relying on prior assumptions that may cause errors, we directly use real finger data. We employ a large gesture database~\cite{Kapitanov_2024_WACV} and the Segment Anything Model~\cite{kirillov2023segany} to segment finger areas and create annotations for the finger ROI, joint cavities, and shapes, ensuring each vein identity has a unique shape. 

To improve the realism of vein image generation, we propose a cascaded region-aware supervision strategy, with the objective function defined as:
\begin{equation}
\small
	\mathcal{L} = \mathcal{L}_{\mathit{cycle}\mathit{GAN}} + \mathcal{L}_{\mathit{region-aware}},
\end{equation}
where \( \mathcal{L}_{\mathit{cycleGAN}} \) includes identity, cycle consistency, and adversarial losses~\cite{zhu2017unpaired}. The region-aware loss \( \mathcal{L}_{\text{Region-Aware}} \) is defined as:
\begin{equation}
	\small
	\mathcal{L}_{\mathit{region-aware}} = \mathcal{L}_{\mathit{cycle}}^s + \mathcal{L}_{\mathit{cycle}}^{\mathit{roi}} + \mathcal{L}_{\mathit{adv}}^s + \mathcal{L}_{\mathit{adv}}^{\mathit{roi}} + \lambda_1 \mathcal{L}_{\mathit{shape}} + \lambda_2 \mathcal{L}_j + \lambda_3 \mathcal{L}_p,
\end{equation}
where  \( \mathcal{L}_{\mathit{cycle}}^s \) and \( \mathcal{L}_{\mathit{cycle}}^{\mathit{roi}} \)  are cycle consistency losses applied to the finger shape region and ROI, while \( \mathcal{L}_{\mathit{adv}}^s \) and \( \mathcal{L}_{\mathit{adv}}^{\mathit{roi}} \) are their corresponding adversarial losses. These losses guide the generator toward producing photorealistic vein images. Discriminators for the finger shape and ROI operate independently of the main discriminator.

 The shape regularization term \( \mathcal{L}_{\mathit{shape}} \) employs Haber loss on binary finger shape masks for anatomical accuracy. The perceptual terms \( \mathcal{L}_j \) and \( \mathcal{L}_p \) enforce brightness priors for joint cavities and vein patterns, respectively:
\begin{equation}
	\mathcal{L} = \frac{1}{N} \sum_{i=1}^N \max\left(0, \mathit{margin} - \left(\mathit{Mean}_\mathcal{A}(x_i) - \mathit{Mean}_\mathcal{B}(x_i)\right)\right),
\end{equation}
where \( x_i \) denotes a synthesized vein image, and \( \mathit{Mean}_{\mathcal{A}/\mathcal{B}} \) compute average intensities over specific regions. This loss reflects the typical intensity distribution in real finger vein images: joint cavities tend to be brighter than surrounding tissue, and vein patterns appear darker. Based on empirical observations, we set \( \mathit{margin}_j = 0.1 \) and \( \mathit{margin}_p = 0.3 \) to enforce these physiological constraints.

\subsection{Intra-Class Diversity Enhancement}
Public finger vein datasets typically exhibit limited intra-class diversity, with minimal variation among samples from the same finger. This constraint limits their effectiveness for unconstrained, non-contact recognition systems, which often operate in open-set scenarios with significant inter-class variation. To address this, we generate a large set of samples for each unique vein identity, introducing diverse variations such as horizontal displacement, rotation, rolling, scaling, under-/overexposure, motion blur, optical blur, skin scattering blur~\cite{nayar1999vision}, and their combinations. Each sample is accompanied by detailed annotations describing its intra-class variations.

\section{The FingerVeinSyn-5M Dataset}
FingerVeinSyn-5M is currently the largest synthetic finger vein image dataset. Additionally, it is the first dataset to provide comprehensive annotations for finger vein images. It includes 50,000 unique finger vein identities, with 100 samples per finger, totaling 5 million samples. Each unique finger has rich intra-class variations, with the 100 samples including 5 horizontal displacements (±20), 5 scalings (±0.15), 9 rollings (±20), 11 rotations (±20), and 5 samples each of under/overexposure, motion blur, optical blur, and skin scattering blur. The final 50 samples include different combinations of these intra-class variations.

FingerVeinSyn-5M uses XML to save annotation information, including the coordinates of finger joint cavities, the center coordinates of the finger ROI bounding boxes, and all intra-class variation information relative to the center view, including shift distances, rotation angles, and scaling factors. Additionally, the dataset provides corresponding binary masks for finger shape positions and finger vein patterns.

\begin{table}[]
	\centering
	\caption{Dataset Evaluation of Synthetic Finger Vein}
	\label{tb2}
	\begin{tabular}{lccc}
		\toprule
		Dataset          & $U_{class}$ & $C_{intra}$ & $D_{intra}$ \\ \midrule
		FingerVeinSyn-5M & 99.83  & 99.99  & 68.30  \\
		{HKPU-SynFV~\cite{hillerstrom2014generating}}         & 19.19  & 99.98  &  3.279 \\ \bottomrule
	\end{tabular}
\end{table}

\begin{table*}[t]
	\centering
	\caption{Performance comparison of real vs. synthetic finger vein data (real data: a combination of six real datasets, synthetic data: FingerVeinSyn-5M).}
	\label{tb3}
	\resizebox{\linewidth}{!}{
	\begin{tabular}{cccccccccccc}
		\toprule
		\multirow{2}{*}{Method} & \multicolumn{4}{c}{Configs}            & \multicolumn{7}{c}{Performance (TAR@FAR=1e-6) $\uparrow$}                                                                              \\ \cmidrule{2-12} 
		& \#IDs & \#per ID & \#Image & FT w/Real & UTFVP           & FV-USM          & HKPU-FV         & SDUMLA-FV       & MMCBNU    & PLUS-FV3        & Avg.            \\ \midrule
	Real data                & 1.38k & -        & 12k     & \XSolidBrush         & 0.1963          & 0.4060          & 0.2685          & 0.5245          & 0.6094          & 0.2701          & 0.3622          \\
	Syn. data~\cite{hillerstrom2014generating}                & 1.6k  & 10       & 16k     & \XSolidBrush         & 0.0472          & 0.0573          & 0.0617         & 0.3174          & 0.5046          & 0.1754          & 0.1939          \\
	Syn. data                & 1.2k  & 10       & 12k     & \XSolidBrush        & 0.7306          & 0.4571          & 0.3567          & 0.5996          & 0.6693          & 0.2703          & 0.5139         \\
	Syn. data                & 1.2k  & 50       & 60k     & \XSolidBrush        & \textbf{0.7824} & 0.5902          & 0.6391          & 0.6851          & 0.8276          & 0.2582          & 0.6304          \\
	Syn. data                & 1.2k  & 100      & 120k    & \XSolidBrush        & 0.7324          & \textbf{0.6831} & \textbf{0.6443} & \textbf{0.7310} & \textbf{0.8450} & \textbf{0.2861} & \textbf{0.6537} \\ \midrule
		Syn. data~\cite{hillerstrom2014generating}                & 1.6k  & 10       & 16k     & \CheckmarkBold         & 0.5880         & 0.6264          & 0.4246          & 0.4998          & 0.8327          & 0.2562          & 0.5380          \\
		Syn. data                & 1.2k  & 10       & 12k     & \CheckmarkBold         & \textbf{0.7703} & \textbf{0.7403} & \textbf{0.7576} & \textbf{0.6509} & \textbf{0.8599} & \textbf{0.4531} & \textbf{0.7054} \\ \midrule
	Syn. data                & 10k   & 10       & 100k    & \CheckmarkBold         & 0.8037          & 0.7919          & 0.7805          & 0.7359          & 0.8899          & 0.4807          & 0.7471          \\
	Syn. data                & 20k   & 10       & 200k    & \CheckmarkBold         & 0.8185          & 0.8466          & 0.7957          & 0.7713          & 0.9244          & 0.7876          & 0.8240          \\
	Syn. data                & 30k   & 10       & 300k    & \CheckmarkBold         & 0.8185          & 0.8481          & 0.7957          & 0.7713          & 0.9243          & \textbf{0.8708} & 0.8381          \\
	Syn. data                & 40k   & 10       & 400k    & \CheckmarkBold         & 0.8046          & \textbf{0.9235} & 0.9185          & 0.8788          & 0.9436          & 0.7072          & 0.8627          \\
	Syn. data                & 50k   & 10       & 500k    & \CheckmarkBold         & \textbf{0.9074} & 0.8798          & \textbf{0.9187} & \textbf{0.9377} & \textbf{0.9463} & 0.8177          & \textbf{0.9013} \\ \bottomrule
	\end{tabular}}
\end{table*}

\section{Experiments}
\subsection{\textbf{Experimental Setup}}
We conducted experiments by dividing each publicly available dataset into training and test sets at a 1:1 ratio with no overlapping identities. We used the TAR (True Accept Rate) @ FAR (False Accept Rate) metric to evaluate the recognition model's performance. Additionally, we assessed the uniqueness, inter-class consistency, and intra-class diversity~\cite{10204758} of our proposed synthetic dataset FingerVeinSyn-5M.

\subsubsection*{\textbf{Datasets}}
We utilized six publicly available datasets: UTFVP~\cite{Ton_UTFVPDB2013}, FV-USM~\cite{asaari2014fusion}, HKPU-FV~\cite{kumar2011human}, SDUMLA-FV~\cite{yin2011sdumla}, MMCBNU\_6000~\cite{lu2013available}, and PLUS-FV3~\cite{8698588}, totaling 2,760 identities and 23,852 finger vein samples. During finger vein sample generation, since not all datasets contained complete finger regions, only HKPU-FV, FV-USM, and PLUS-FV3 were used for region supervision and shape supervision.

\subsubsection*{\textbf{Generator Model}}
We generated 50,000 identities, each with 100 samples. The synthetic vein images had dimensions of $300\times600$, while the finger-shaped vein images and finger vein ROI images were sized at $200\times600$. We employed the Adam optimizer ($\beta_1=0.5$, $\beta_2=0.999$) for model training, with an initial learning rate of $2e-4$ that decayed after 50 epochs. The generator model was implemented using the PyTorch framework and trained on 4 NVIDIA RTX 4090 GPUs with a batch size of 24.

\subsubsection*{\textbf{Recognition Model}}
The recognition model utilized a ResNet101 backbone~\cite{he2016deep} integrated with ArcFace~\cite{deng2019arcface} (scale factor $s=32$, margin $M=0.5$). Both training and fine-tuning were performed for 20 epochs on finger vein ROI images sized at $200 \times 600$, with a learning rate of $0.1$ for training and $1e-4$ for fine-tuning, respectively. The model was implemented in PyTorch and trained on 4 NVIDIA RTX 4090 GPUs with a batch size of 128.

\subsection{Dataset Evaluation}
In this section, we follow the three dependency-type metrics~\cite{10204758} to evaluating synthetic datasets to help us understand the properties of the generated datasets. \( F_{eval} \) is the recognition model used to evaluate synthetic finger vein datasets. The more generalizable \( F_{eval} \) is, the more accurate the metrics are in terms of the identity and diversity of the generated synthetic datasets. Let \( y_c \) be the class label, and \( f_i = F_{eval}(x_i) \). \( d(f_i, f_j) \) represents the distance between two images in the \( F_{\text{eval}} \) feature space.
\subsubsection*{\textbf{Uniqueness}}
The Uniqueness \(U_{class} \) can be used to quantify the number of distinct identities generated in a synthetic dataset. During evaluation, we determine the capacity limit of the synthetic dataset \(U\) by constructing non-overlapping r-ball regions in the \(F_{eval}\) feature space—continuing to add new regions until no further additions can be made without overlap. This process is constrained by two key parameters: the feature distance threshold r (used to determine identity matching) and the feature extraction capability of the \(F_{eval}\) model.

In implementation, we first compute the mean feature vector (cluster center) for all samples belonging to the same identity. If the distance between two cluster centers exceeds threshold \( r \), they are classified as distinct identities. The uniqueness metric \( U_{\text{class}} \) is then defined as \( \frac{|U_c|}{C} \), where the numerator \( |U_c| \) represents the actual count of unique identities generated in the synthetic dataset, and the denominator \( C \) denotes the total number of predefined identity classes in \( F_{eval} \).
As shown in Table \ref{tb2}, the identity distinctiveness of ‌FingerVeinSyn-5M‌ is ‌99.83\%‌, indicating that the generated finger vein patterns exhibit significant differences. Only a very small portion of the finger vein identity feature vectors show low similarity ($r < 0.2$).
\subsubsection*{\textbf{Intra-Class Consistency}}
\(C_{intra}\) is used to evaluate feature coherence among samples belonging to the same unique identity, quantified by measuring the distance between each sample and its corresponding identity feature center. This experiment uses the bounding box annotation provided by the synthetic dataset to extract finger vein ROI images for recognition, benefiting from the accurate annotation, the FingerVeinSyn-5M dataset has an intra-class consistency $C_{intra}$ of 99.99\%.
\subsubsection*{\textbf{Intra-Class Diversity}}
‌\(D_{intra}\) measures the variation among generated samples within the same identity. Given the inherently limited intra-class variations in existing public finger vein datasets, we cannot directly adopt the method from~\cite{10204758}. Instead, we train a recognition model \(F_{\mathit{n}}\) using aligned and normalized samples. In this feature space, two samples from the same identity are considered distinct if their distance exceeds threshold r. The metric is formally defined as
\begin{equation}
D_{\text{intra}} = \left( \frac{1}{C} \right)\left( \frac{2}{N(N-1)} \right) \times \sum_{c=1}^{C} \sum_{i=1}^{N} \sum_{j=i+1}^{N} \delta[d({f_{\mathit{n}}}_i^{c}, {f_{\mathit{n}}}_j^{c}) < r]
\end{equation}
where \( f_{n}^{c} \) is the \( n \)-th sample of class \( c \), and \( \delta[\text{condition}] = 1 \) if the condition is true, otherwise it is 0. \( D_{\text{intra}} \) ranges from 0 to 1, with higher values indicating greater intra-class variation. 

As shown in Table \ref{tb2}, the intra-class diversity of FingerVeinSyn-5M is 68.3\%. This is due to the fact that the generated intraclass variations include both geometric and optical variations, where optical variations are necessary, but it produces relatively weak interclass variations. The interclass diversity of geometric changes alone is as high as 98.3\%
\begin{figure*}[t]
	\captionsetup[subfloat]{labelsep=none,format=plain,labelformat=empty}
	\centering 
	
	\subfloat[]{
		\includegraphics[width=0.166\linewidth]{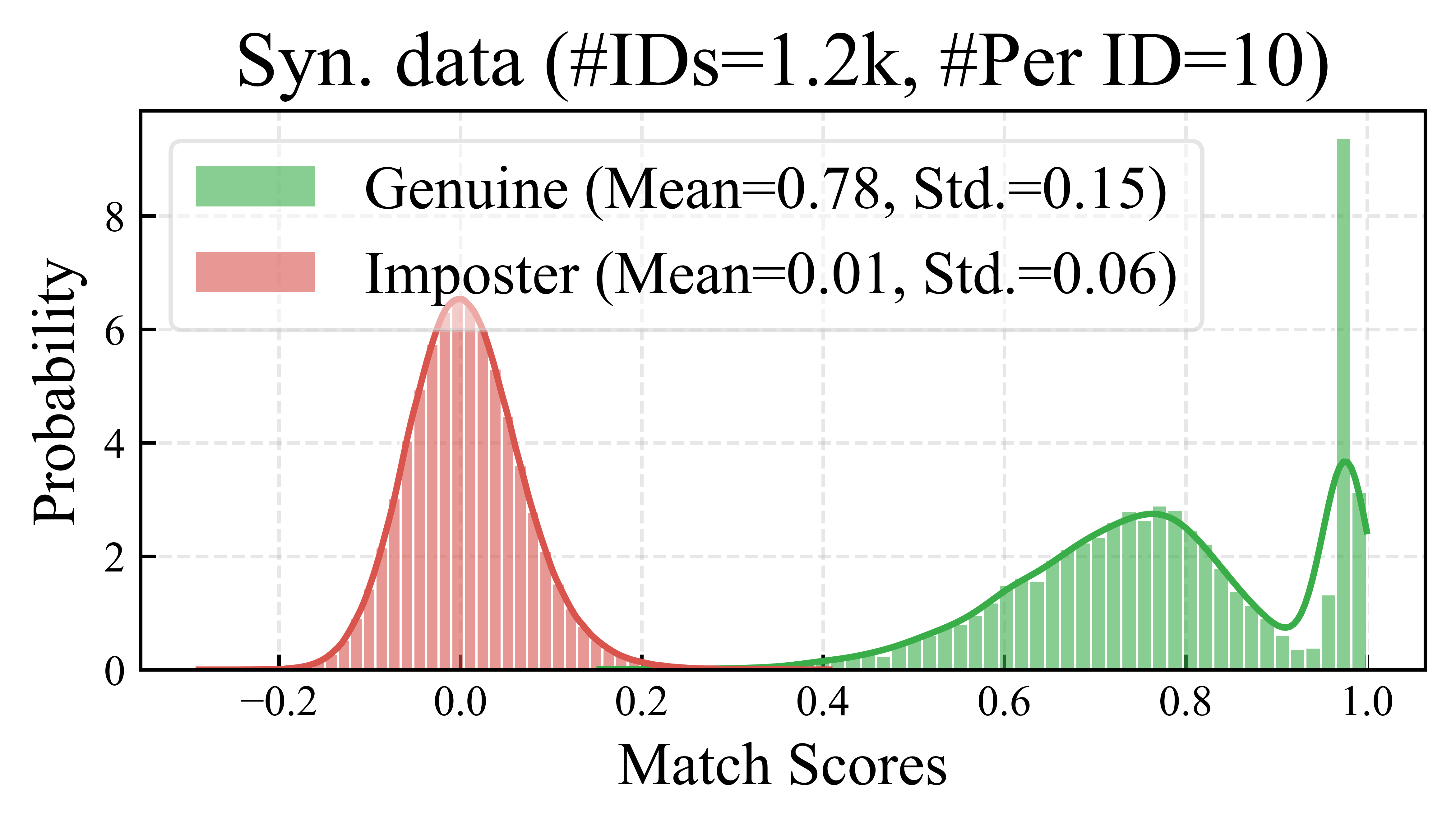}}
	\hspace*{-0.01\linewidth} 
	\subfloat[]{
		\includegraphics[width=0.166\linewidth]{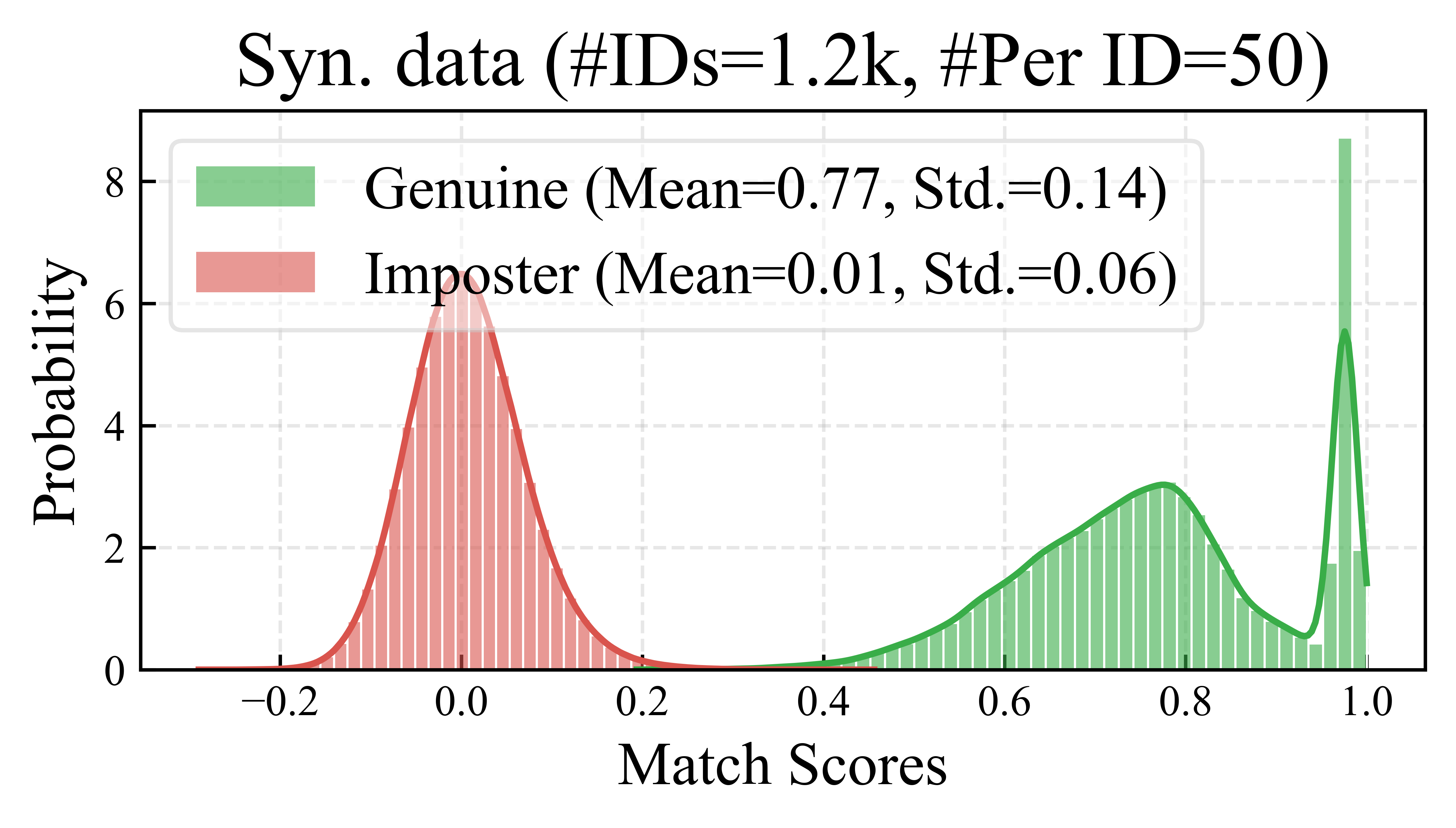}}
	\hspace*{-0.01\linewidth}
	\subfloat[]{
		\includegraphics[width=0.166\linewidth]{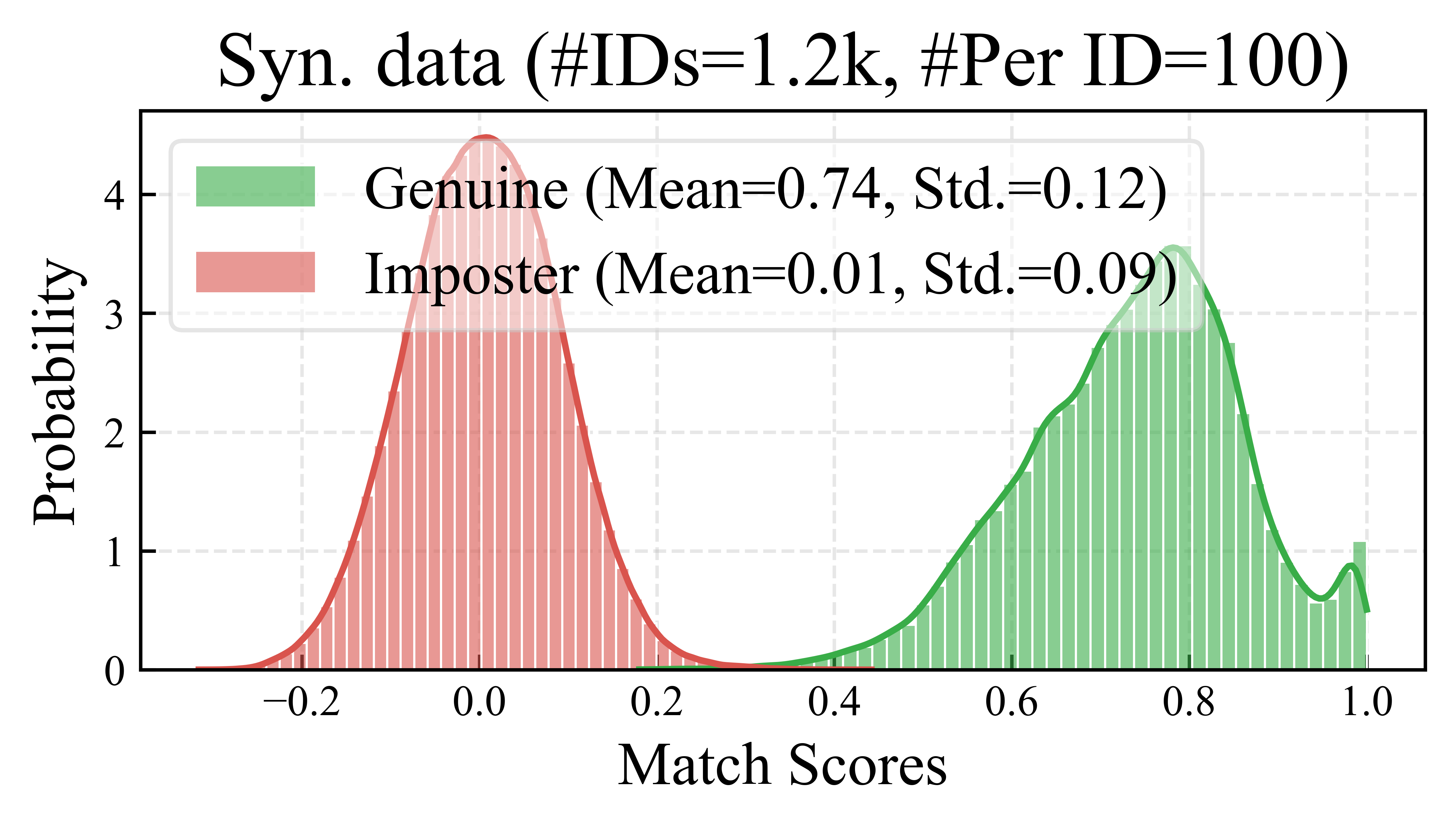}}
	\hspace*{-0.01\linewidth}
	\subfloat[]{
		\includegraphics[width=0.166\linewidth]{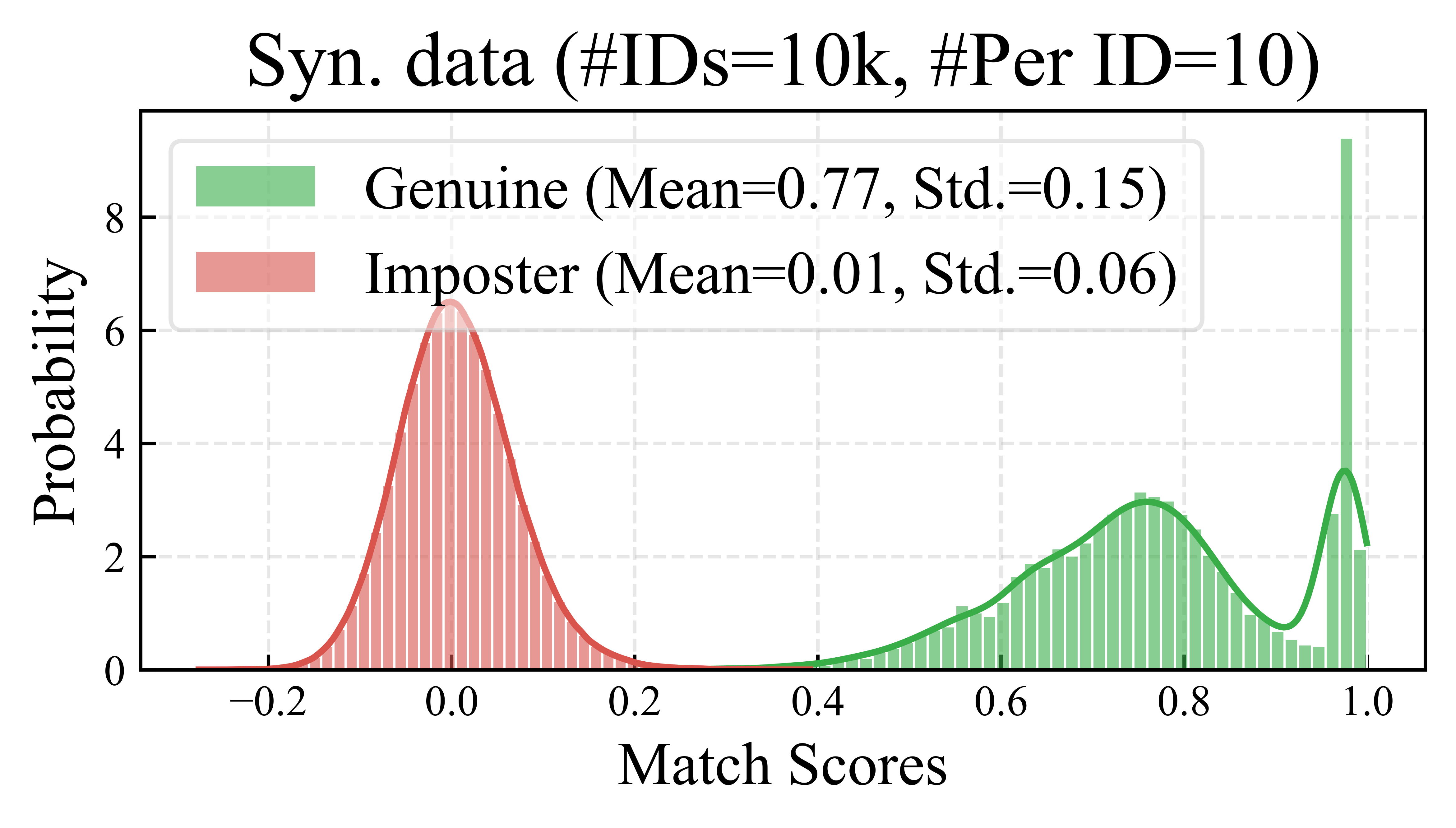}}
	\hspace*{-0.01\linewidth}
	\subfloat[]{
		\includegraphics[width=0.166\linewidth]{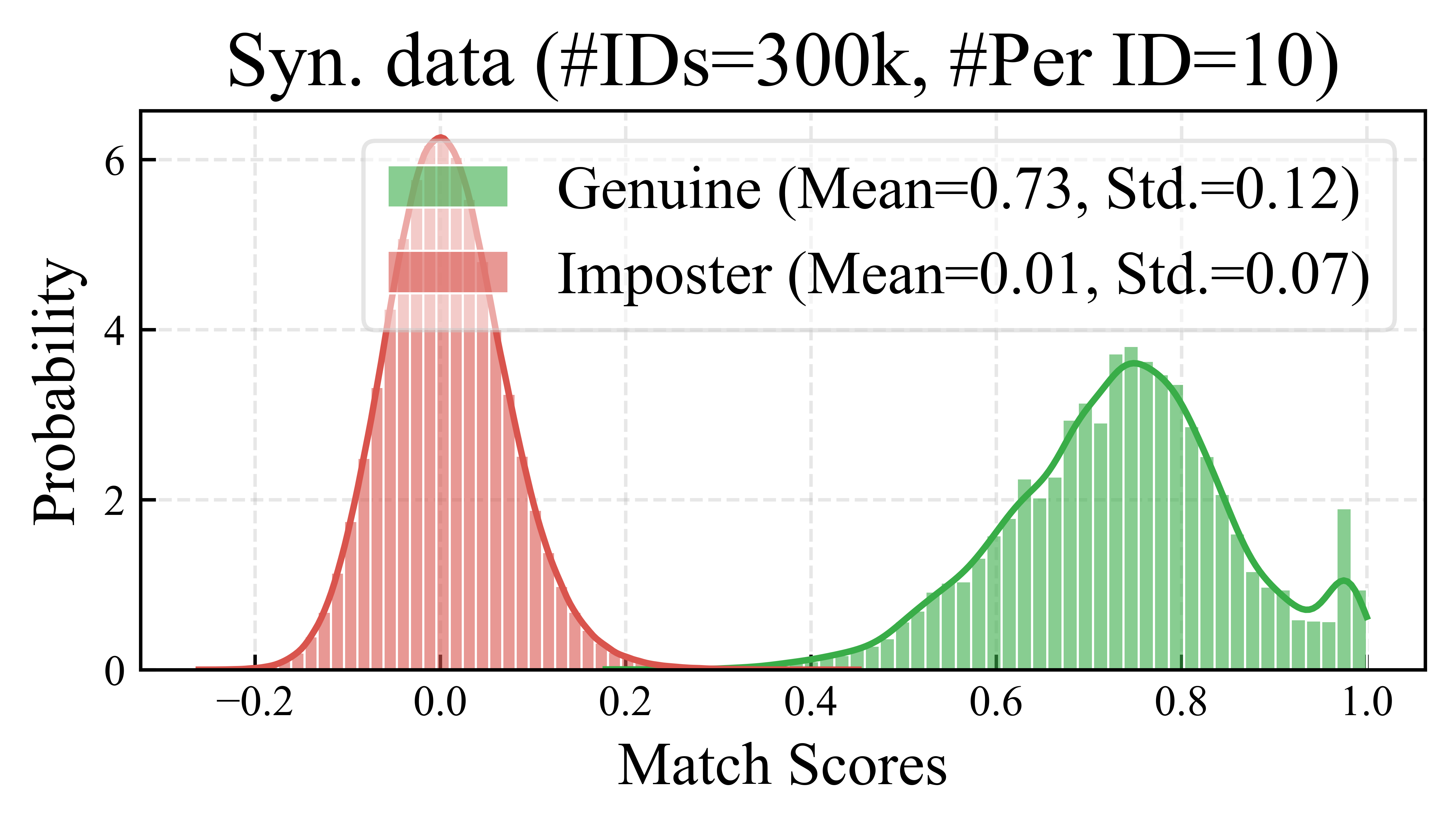}}
	\hspace*{-0.01\linewidth}
	\subfloat[]{
		\includegraphics[width=0.166\linewidth]{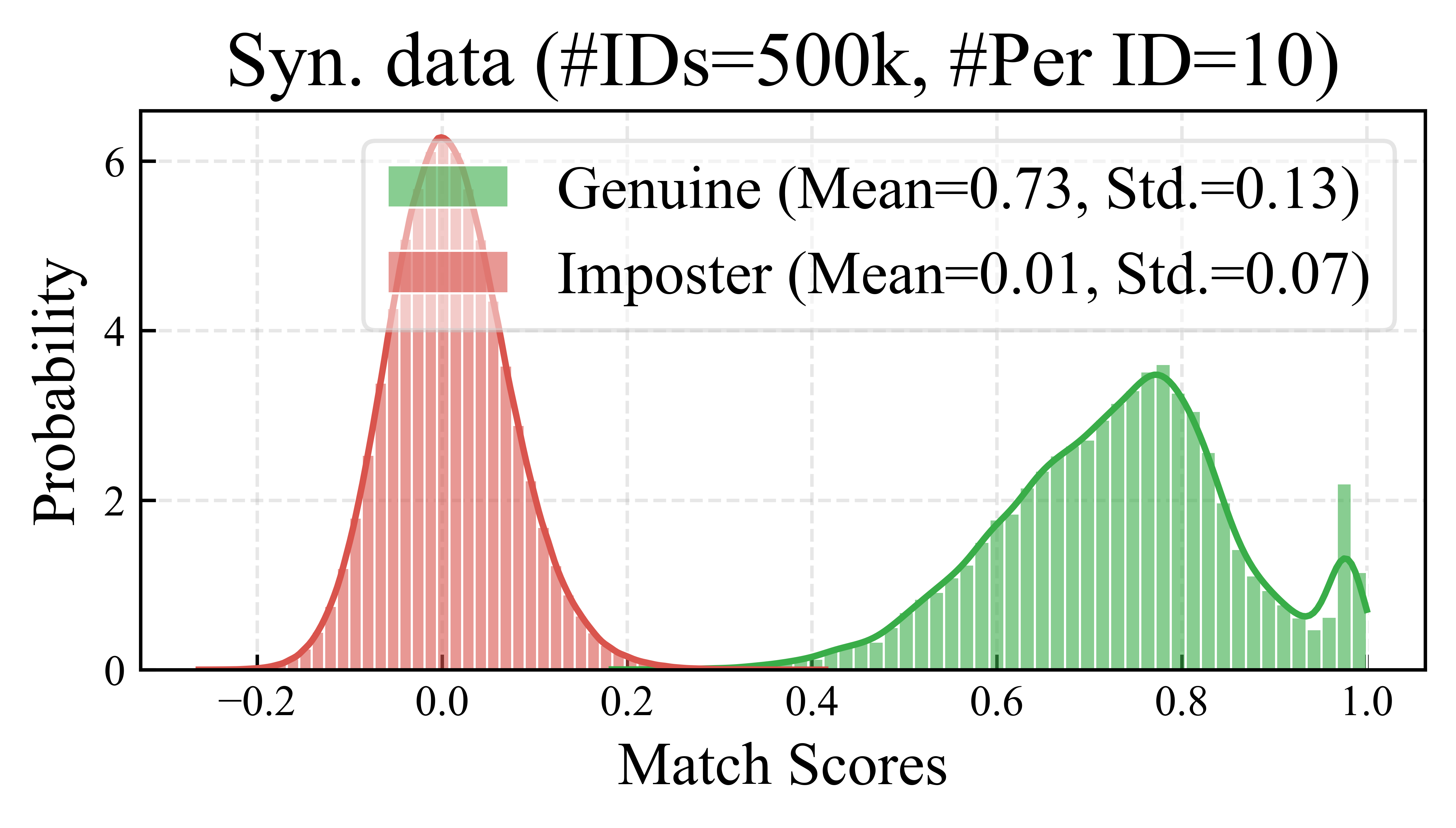}}
	
	\vspace{-20pt} 
	
	\subfloat[]{
		\includegraphics[width=0.166\linewidth]{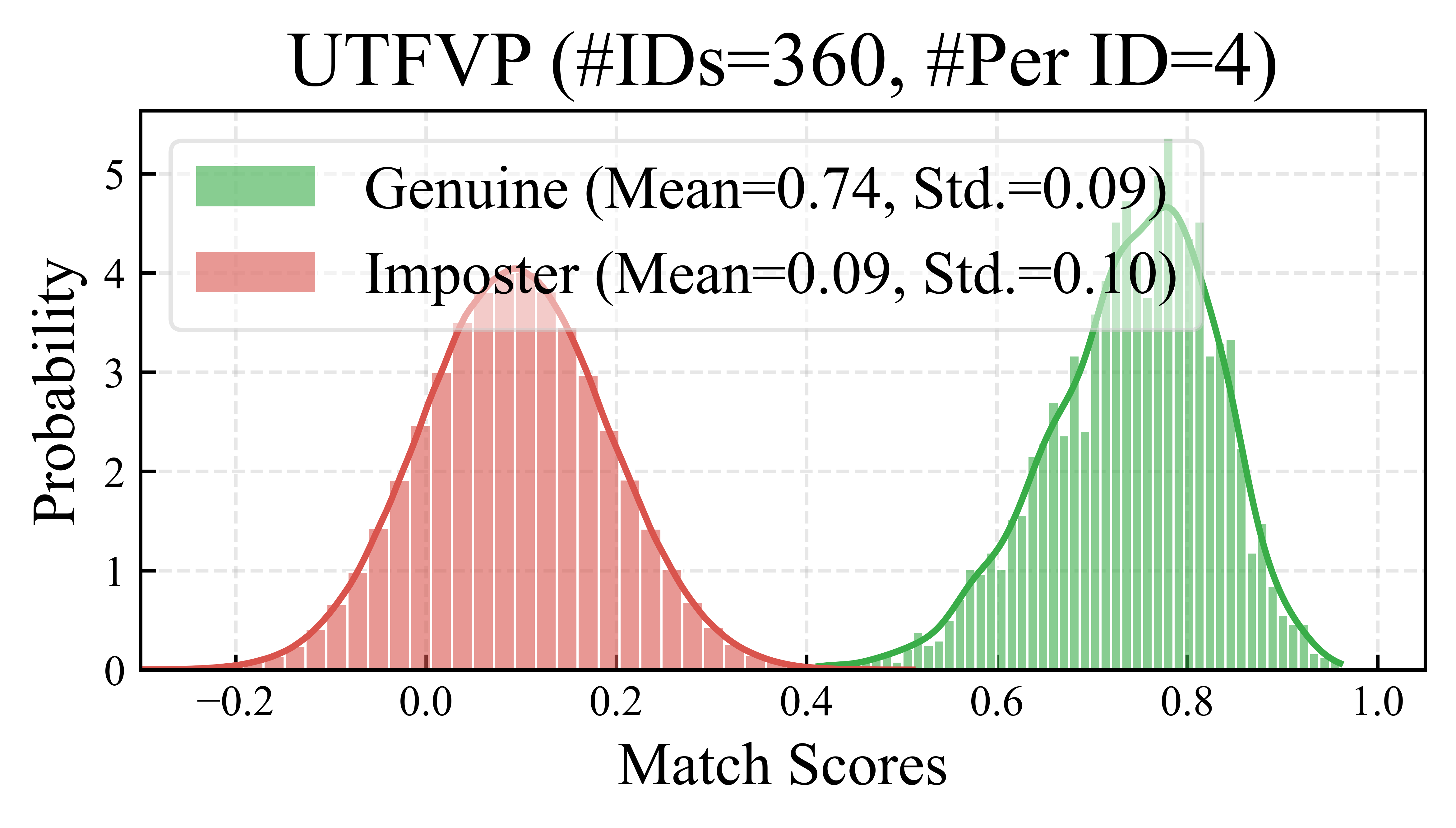}}
	\hspace*{-0.01\linewidth}
	\subfloat[]{
		\includegraphics[width=0.166\linewidth]{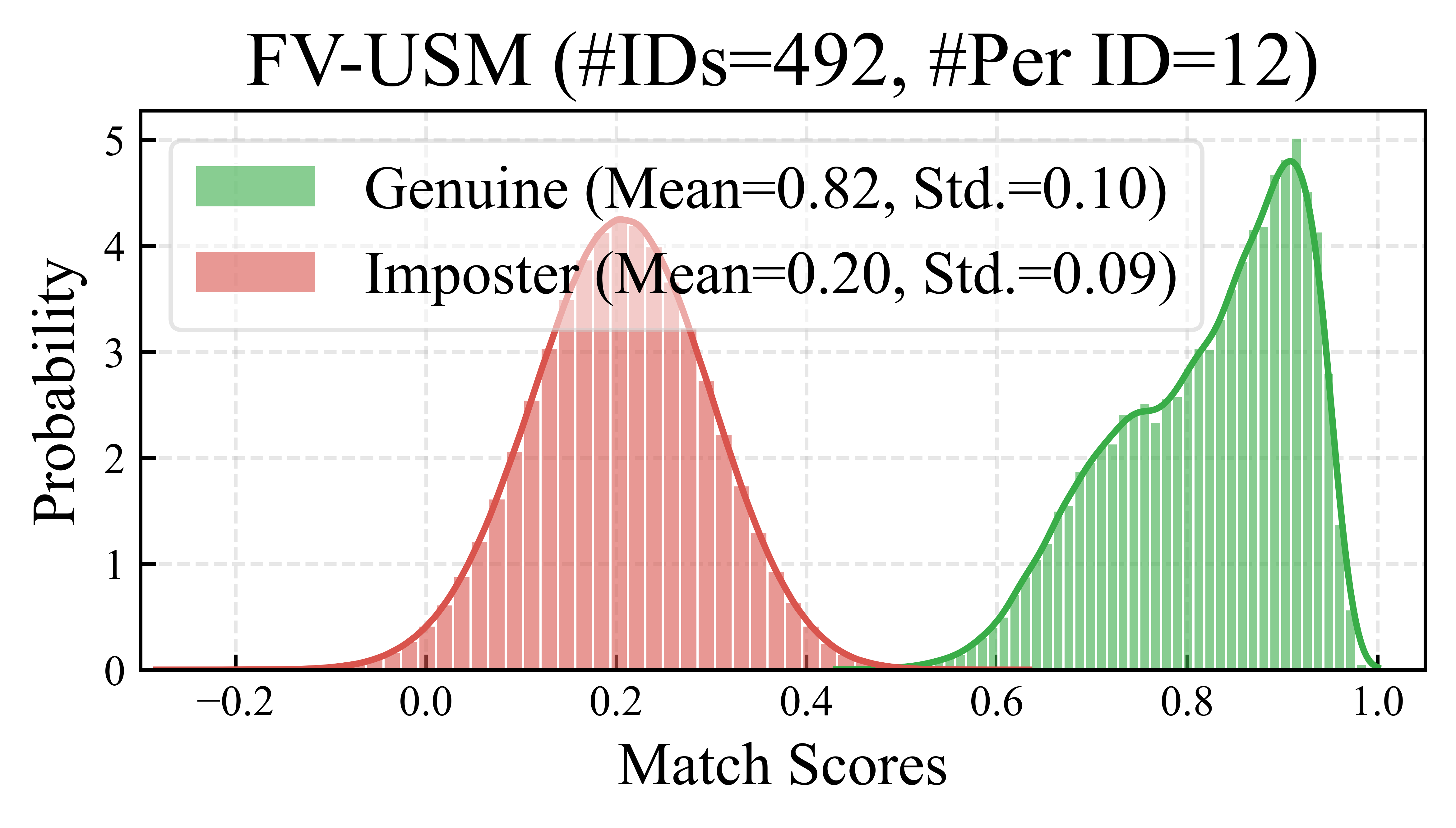}}
	\hspace*{-0.01\linewidth}
	\subfloat[]{
		\includegraphics[width=0.166\linewidth]{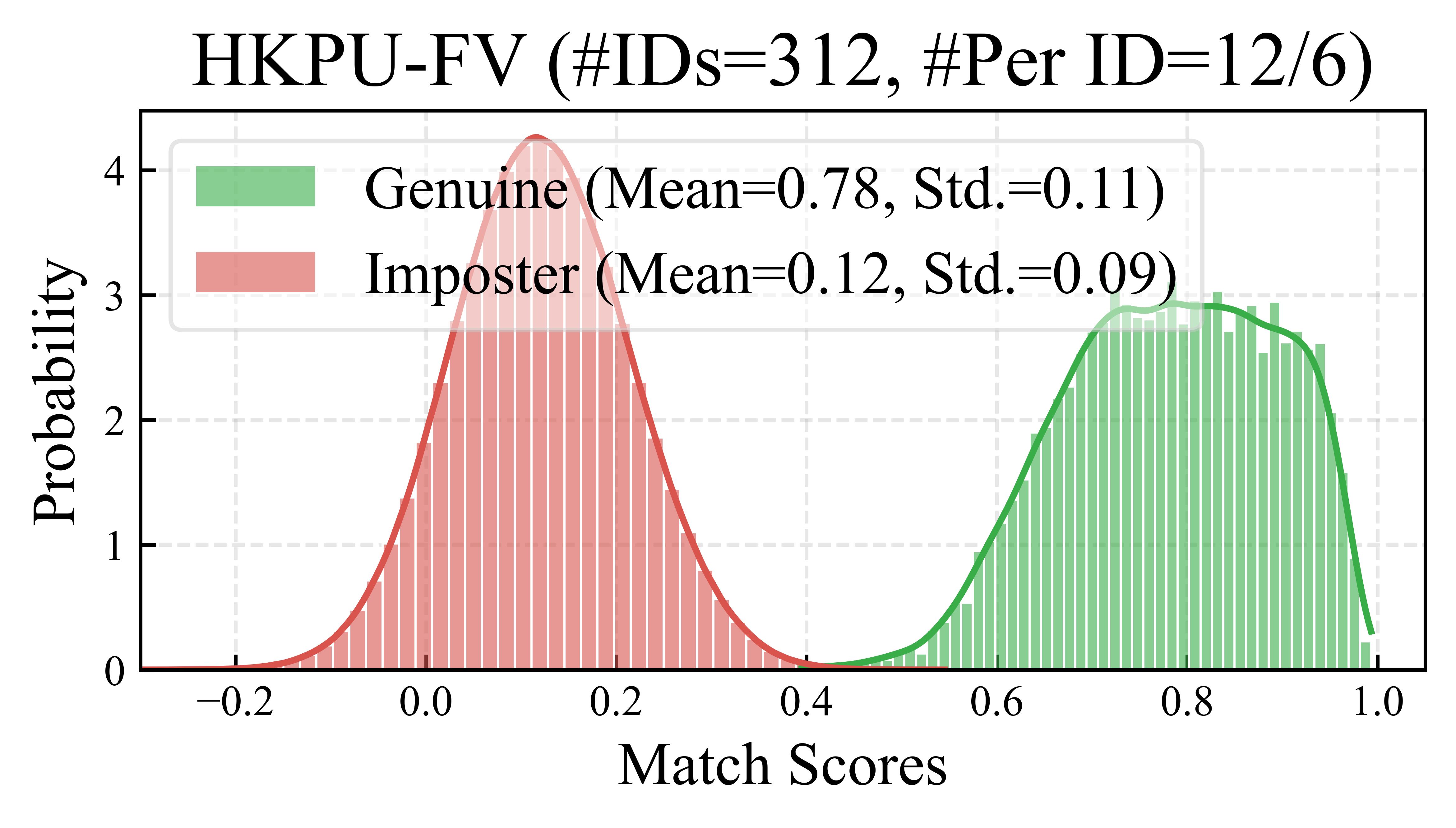}}
	\hspace*{-0.01\linewidth}
	\subfloat[]{
		\includegraphics[width=0.166\linewidth]{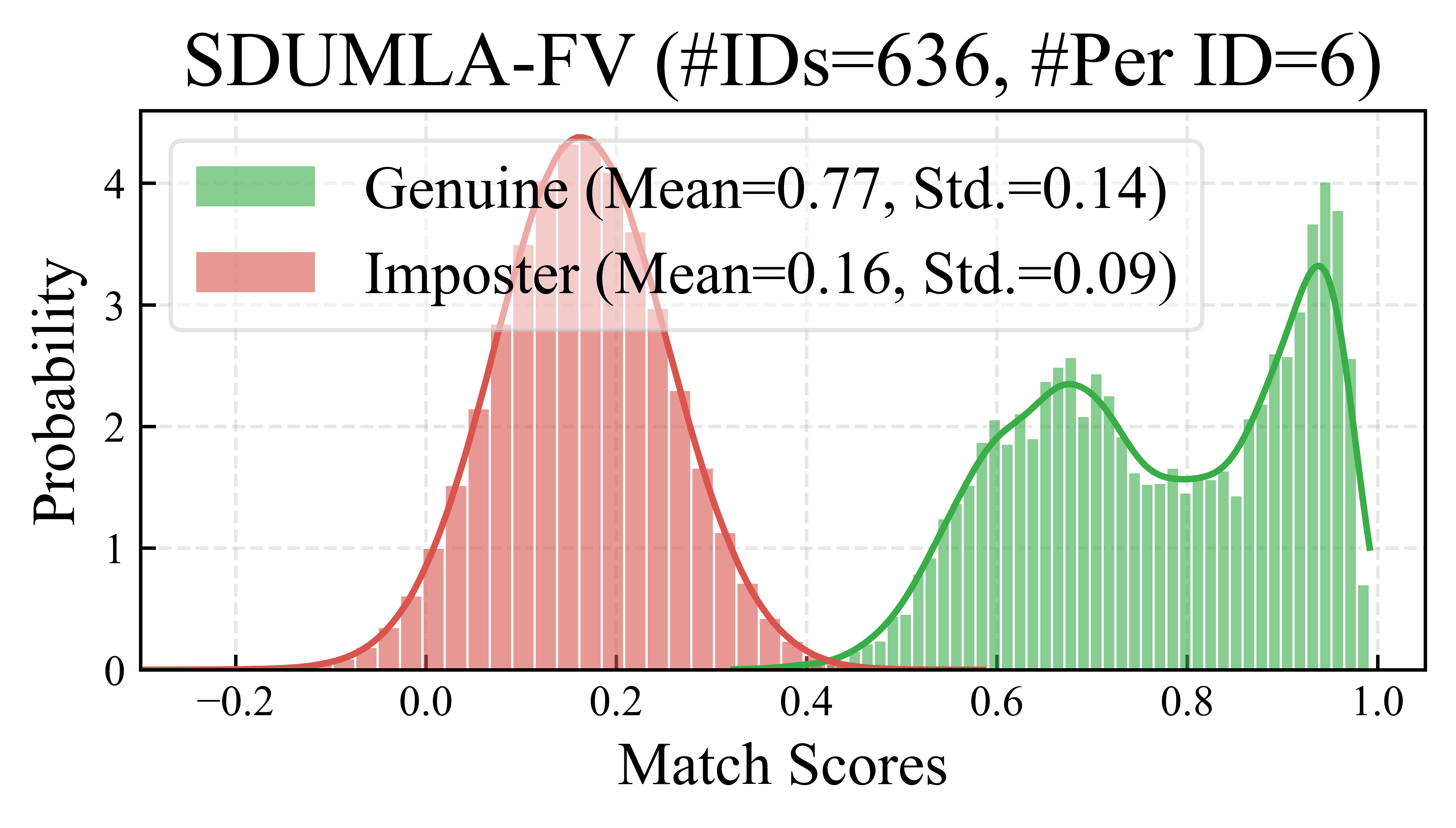}}
	\hspace*{-0.01\linewidth}
	\subfloat[]{
		\includegraphics[width=0.166\linewidth]{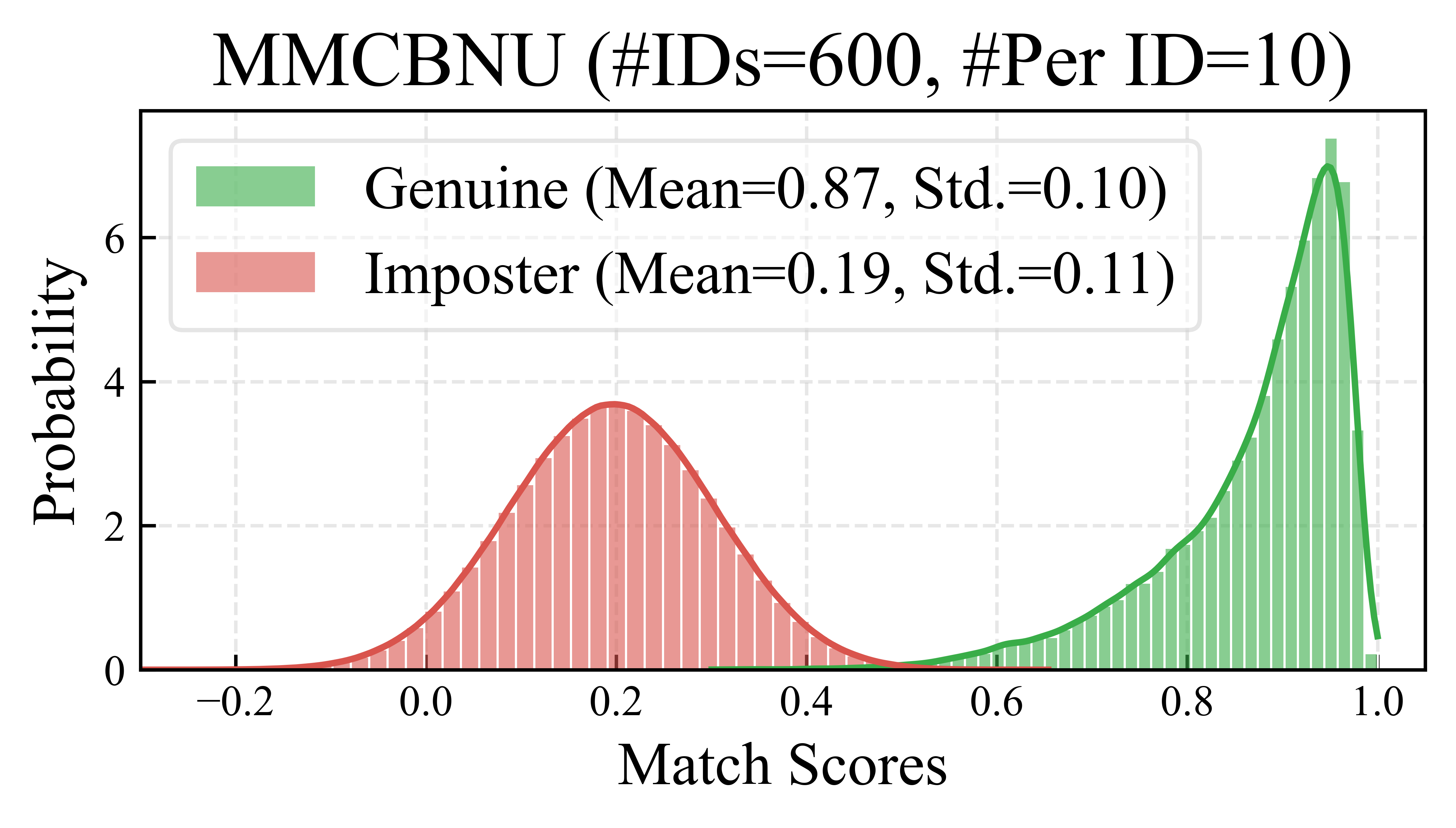}}
	\hspace*{-0.01\linewidth}
	\subfloat[]{
		\includegraphics[width=0.166\linewidth]{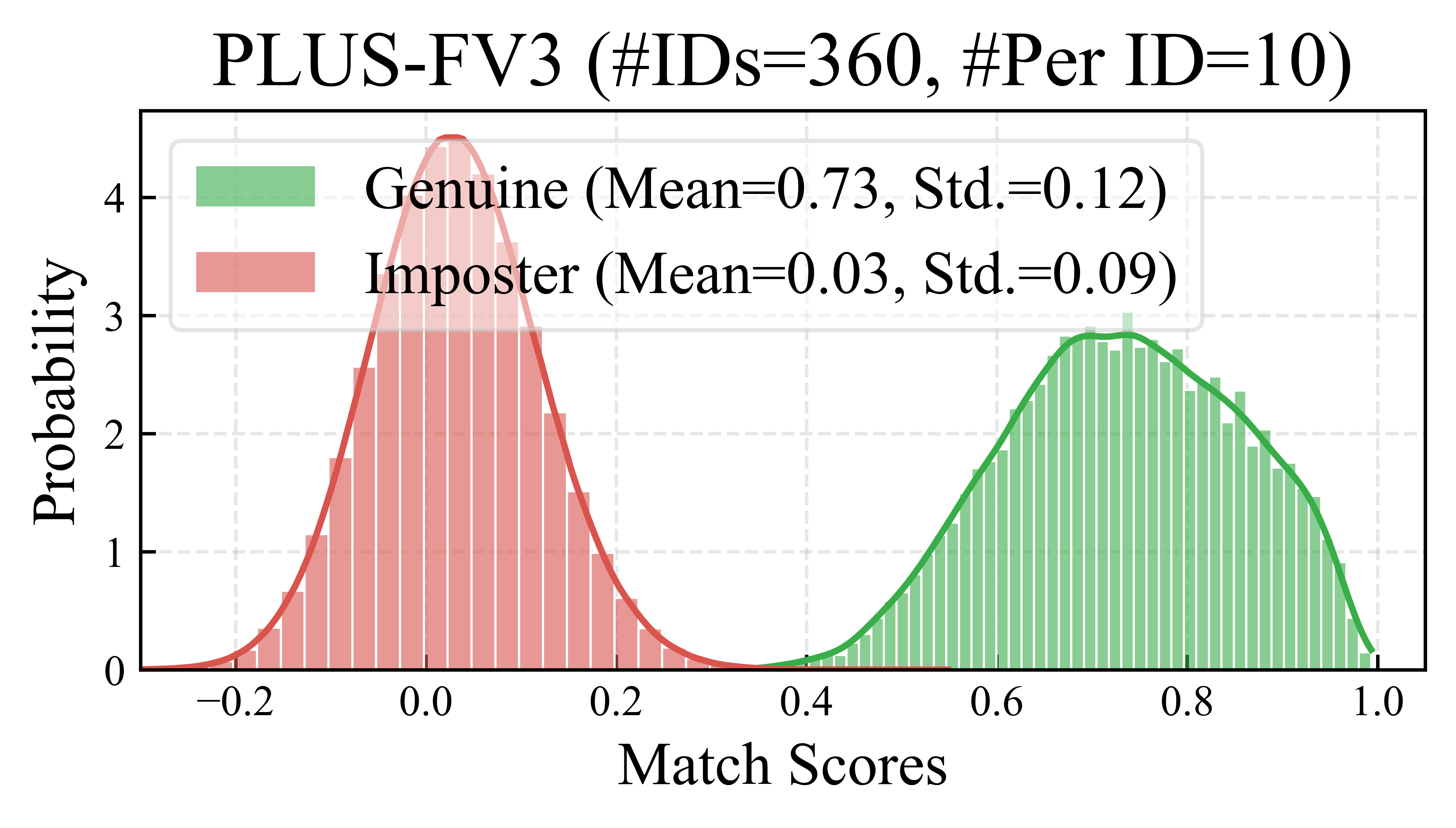}}
	\vspace{-20pt}
	\caption{Illustration of the matching score distributions between real and synthetic finger vein images.}
	\label{fig2} 
\end{figure*}

\subsection{Open-Set Finger Vein Recognition}
In the Open-Set protocol scenario, we compared recognition performance between real and synthetic data. The experimental design adopted non-overlapping identities to ensure complete independence between the training and test set. As shown in ‌Table \ref{tb3}, comparative analysis with other real datasets indicates that ‌FingerVeinSyn-5M synthetic data‌ achieves ‌better performance‌ than real data under the same identity and sample size conditions, with an ‌average improvement of 15.17\%‌. Further increasing the diversity of the training set improved performance by 29.15\%. Additionally, when fine-tuned with real data, the model achieved a ‌further accuracy improvement of 53.91\%‌. These results demonstrate that ‌FingerVeinSyn-5M synthetic data‌ can effectively ‌enhance the generalization ability‌ of finger vein recognition systems.

\subsection{Cross-Domain Finger Vein Recognition}
Cross-domain performance is a critical challenge for finger vein recognition systems in practical applications. We evaluate recognition models trained with real/synthetic data, with their performance subsequently validated on cross-domain datasets. Table \ref{tb4} presents the comparative performance results.
The limited size of real datasets, coupled with the domain bias arising from different imaging conditions, results in suboptimal cross-domain recognition performance. This shortcoming stems from the model's inability to effectively generalize variations in illumination, background, and finger position when trained on a limited number of data samples. FVeinSyn addresses this challenge by synthesizing different interclass variations, which enables the model to obtain a more comprehensive feature representation. This approach enhances the model's generalization to different lighting conditions, environmental backgrounds, and finger variations, thereby improving cross-domain recognition performance.

\begin{table}[]
	\centering
	\caption{Performance comparison of real vs. synthetic finger vein data for cross-domain recognition.}
	\label{tb4}
	\resizebox{\linewidth}{!}{
	\begin{tabular}{ccccccc}
		\toprule
		Dataset   & UTFVP  & FV-USM  & HKPU & SDUMLA & MMCBNU & PLUS \\ \midrule
		UTFVP     & -      & 0.1795  & 0.0250  & 0.2400    & 0.4401 & 0.1301   \\
		FV-USM    & 0.2852 & -       & 0.0250  & 0.4324    & 0.5315 & 0.0830   \\
		HKPU   & 0.3662 & 0.0186  & -       & 0.4333    & 0.5984 & 0.1606   \\
		SDUMLA & 0.2125 & 0.4107  & 0.0274  & -         & 0.1152 & 0.0080    \\
		MMCBNU    & 0.2940 & 0.3059  & 0.0504  & 0.3521    & -      & 0.0028   \\
		PLUS-FV3  & 0.0542 & 0.2813 & 0.0094  & 0.1905    & 0.0904 & -        \\ \midrule
	Syn. data~\cite{hillerstrom2014generating} & 0.0472 & 0.0573  & 0.0617  & 0.3174    & 0.5046 & 0.1754   \\
	Syn. data & \textbf{0.7324} & \textbf{0.6831}  & \textbf{0.6443}  & \textbf{0.7310}    & \textbf{0.8450} & \textbf{0.2861}   \\ \bottomrule
	\end{tabular}}
\end{table}
\subsection{One/Few Shot Finger Vein Recognition}
We consider a more practical application scenario where users provide only a limited number of samples or even a single sample during registration. We evaluated the effect of pre-training using synthetic samples and fine-tuning using $N$ registration samples for fine-tuning $(N = 1, 2, 3)$. We set both the real data training and fine-tuning epochs to 60 in this experiment. Table \ref{tb5} compares the recognition performance between models trained solely on real data and those pre-trained on synthetic data followed by fine-tuning. The experimental results show that it is difficult to achieve accurate recognition by training the recognition system using only a limited number of samples. Instead, using FingerVeinSyn-5M synthetic finger vein dataset pre-training followed by fine-tuning using limited samples can effectively improve the recognition performance.
\begin{table}[]
	\centering
	\caption{Performance comparison of one/few-shot finger vein recognition.}
	\label{tb5}
	\resizebox{\linewidth}{!}{
	\begin{tabular}{cccccccc}
		\toprule
		\multicolumn{2}{c}{Dataset} & UTFVP  &FV-USM & HKPU & SDUMLA & MMCBNU & PLUS \\ \midrule
		\multicolumn{1}{c|}{\multirow{2}{*}{$N=1$}} & Real data & 0.0380 & 0.1375 & 0.0553  & 0.4520    & 0.7072       & 0.0874   \\
		\multicolumn{1}{c|}{}                     & \cellcolor{gray!20}Syn. data & \cellcolor{gray!20}0.6667 & \cellcolor{gray!20}0.4367 & \cellcolor{gray!20}0.5658  & \cellcolor{gray!20}0.6895    & \cellcolor{gray!20}0.8607       & \cellcolor{gray!20}0.3795   \\ \midrule
		\multicolumn{1}{c|}{\multirow{2}{*}{$N=2$}} & Real data & 0.2389 & 0.4560 & 0.2144  & 0.5934    & 0.8207       & 0.3472   \\
		\multicolumn{1}{c|}{}                     & \cellcolor{gray!20}Syn. data & \cellcolor{gray!20}0.9444 & \cellcolor{gray!20}0.9458 & \cellcolor{gray!20}0.7752  & \cellcolor{gray!20}0.8422    & \cellcolor{gray!20}0.9448       & \cellcolor{gray!20}0.6307   \\ \midrule
		\multicolumn{1}{c|}{\multirow{2}{*}{$N=3$}} & Real data & -      & 0.5711 & 0.5934  & 0.7348    & 0.9619      & 0.4491   \\
		\multicolumn{1}{c|}{}                     & \cellcolor{gray!20}Syn. data & \cellcolor{gray!20}-      & \cellcolor{gray!20}0.9488 & \cellcolor{gray!20}0.9527  & \cellcolor{gray!20}0.9172    & \cellcolor{gray!20}0.9748       & \cellcolor{gray!20}0.9189   \\ \midrule
		\end{tabular}}
\end{table}

\subsection{Comparison Between Real and Synthetic Finger Vein Images}
Through a comparative analysis of the matching score distributions between real finger vein images and the FingerVeinSyn-5M synthetic dataset under different identity and sample size conditions (shown in Figure \ref{fig2}), we observed significant statistical consistency between the two. Specifically, in subsets of FingerVeinSyn-5M composed of samples from different identities, the matching score distributions of synthetic data and real data exhibited high agreement in key statistical characteristics such as mean and variance. This validates the reliability of the FingerVeinSyn-5M synthetic dataset in terms of authenticity and diversity. It can be observed that using only limited real data for training fails to achieve satisfactory recognition performance. However, pre-training on the synthetic dataset followed by fine-tuning with limited real data significantly improves recognition performance. The experimental results demonstrate that pre-training with the FingerVeinSyn-5M synthetic dataset effectively addresses the sample scarcity problem in finger vein recognition.

\section{Conclusion}
This paper proposes a synthetic finger vein image generator called FVeinSyn, which is capable of generating samples resembling real finger veins. Based on FVeinSyn, we have released the largest-scale finger vein dataset currently available, FingerVeinSyn-5M, which contains 5 million fully annotated finger vein images from 50,000 different fingers, with 100 samples provided for each finger. We introduced various controllable transformations and blurring effects to generate inter-class variations, including shift, rotation, scaling, rolling, as well as various blurs caused by acquisition environments. Furthermore, this dataset is the first to provide comprehensive annotations related to finger vein images, including finger shape, finger vein ROI bounding boxes, finger joint cavity locations and Inter-class variation. It can effectively support the development of deep learning-based finger vein recognition technologies for multiple downstream tasks.

\begin{acks}
	This work was supported by Postgraduate Research \& Practice Innovation Program of Jiangsu Province (KYCX25\_19).
\end{acks}

\bibliographystyle{ACM-Reference-Format}
\bibliography{sample-base}

\end{document}